%% file: main.tex
\documentclass{vldb}
\usepackage{graphicx}
\usepackage{balance}  
\usepackage{xspace}
\usepackage{url}
\usepackage{cite}
\usepackage{enumitem}
\usepackage{amsmath}
\usepackage{amssymb}
\usepackage{theorem}
\usepackage{amssymb}
\usepackage{caption}
\usepackage{subcaption}
\usepackage{color}
\usepackage{array}
\usepackage{comment}
\usepackage[table, xcdraw]{xcolor}
\usepackage{colortbl,listings}
\usepackage{textcomp}
\usepackage{tikz-qtree}
\usepackage{tikz}
\newsavebox{\mybox}
\usepackage{graphicx}

\newtheorem{theorem}{Theorem}
\newtheorem{definition}[theorem]{Definition}

\makeatletter
\def\blfootnote{\xdef\@thefnmark{}\@footnotetext}
\makeatother

\newcommand{\sparagraph}[1]{\vspace{1mm}\noindent {\bf #1}}

\newcommand{\system}{\textsf{ARDA}\xspace}
\newcommand{\AML}{AML\xspace}

\vldbTitle{Working title}
\vldbAuthors{MIT people}
\vldbDOI{https://doi.org/10.14778/xxxxxxx.xxxxxxx}
\vldbVolume{12}
\vldbNumber{xxx}
\vldbYear{2019}

\usepackage[utf8]{inputenc}
\usepackage{amsmath,amssymb}
\usepackage{cite}
\usepackage{url}
\usepackage{multirow}
\newcommand{\Tau}{\mathrm{T}}

\usepackage[framemethod=TikZ]{mdframed}
\newcounter{Frame}
\newenvironment{Frame}[1][htb]{%
\refstepcounter{Frame}
    \begin{mdframed}[%
        frametitle={#1},
        skipabove=\baselineskip plus 2pt minus 1pt,
        skipbelow=\baselineskip plus 2pt minus 1pt,
        linewidth=1.0pt,
        frametitlerule=true,
        nobreak=true
    ]%
}{%
    \end{mdframed}
}
\usepackage{multirow}

\begin{document}


\title{ARDA: Automatic Relational Data Augmentation for Machine Learning}



%
%
%
%

\numberofauthors{1} 

\author{
  \alignauthor
  Nadiia Chepurko$^1$, Ryan Marcus$^1$, Emanuel Zgraggen$^1$, \\ Raul Castro Fernandez$^2$, Tim Kraska$^1$, David Karger$^1$ \\
  \affaddr{$^1$MIT CSAIL \quad $^2$University of Chicago}\\
  \email{\{nadiia, ryanmarcus, emzg, kraska, karger\}@csail.mit.edu \quad raulcf@uchicago.edu}
\\\
}

\maketitle

\begin{abstract}
Automatic machine learning (\AML) is a family of techniques to automate the process of training predictive models, aiming to both improve performance and make machine learning more accessible. While many recent works have focused on aspects of the machine learning pipeline like model selection, hyperparameter tuning, and feature selection, relatively few works have focused on automatic data augmentation. Automatic data augmentation involves finding new features relevant to the user's predictive task with minimal ``human-in-the-loop'' involvement.


We present \system, an end-to-end system that takes as input a dataset and a data repository, and outputs an augmented data set such that training a predictive model on this augmented dataset results in improved performance. Our system has two distinct components: (1) a framework to search and join data with the input data, based on various attributes of the input, and (2) an efficient feature selection algorithm that prunes out noisy or irrelevant features from the resulting join. We perform an extensive empirical evaluation of different system components and benchmark our feature selection algorithm on real-world datasets.
\end{abstract}

\input{intro_alternative}

\input{system_model}
\input{joins}
\input{techniques}

\input{experiments}

\input{related_work}
\input{conclusions}

\clearpage

\bibliographystyle{abbrv}
\bibliography{lit,ryan-cites-long} \



\end{document}

%% file: intro_alternative.tex
\section{Introduction}
\label{sec:intro}

Automatic machine learning (\AML) aims to significantly simplify the process of building predictive models. 
In its simplest form, \AML tools automatically try to find the best machine learning algorithm and hyper-parameters for a given prediction task~\cite{thornton2013auto, sparks2015tupaq, li2017hyperband , 10.1145/3299869.3319863}. 
With \AML, the user only has to (1) provide a dataset (usually a table) with features and a predictive target (i.e., columns), and (2) specify their model goal (e.g., build a classifier and maximize the F1 score). 
More advanced \AML tools go even further and not only try to find the best model and tune hyper-parameters, but also perform automatic feature engineering. 
In their current form, \AML tools can  outperform experts~\cite{he2019automl} and can help to make ML more accessible to a broader range of users \cite{northstar,fusi2018probabilistic}. 

However, what if the original dataset provided by the user does not contain enough signal (predictive features) to create an accurate model? 
For instance, consider a user who wants to use the publicly-available NYC taxi dataset~\cite{illinoisdatabankIDB9610843} to build a forecasting model for taxi ride durations. 
Suppose the dataset contains trip information over the last five years, including license plate numbers, pickup locations, destinations, and pickup times. 
A model built only on this data may not be very accurate because there are other major external factors that impact the duration of a taxi ride.
For example, weather obviously has a strong impact on the demand of taxis in NYC, and large events like baseball games can significantly alter traffic. 

While it is relatively easy to find related datasets using systems like Google Data Search \cite{brickley2019google} or standardized repository like Amazon's data collections \cite{paolacci2010running}, trying to integrate these datasets as part of the feature engineering process remains a challenge. First, integrating related datasets requires finding the right join key between different datasets. 
Consider, for example, the hypothetical taxi dataset and a related NYC weather dataset. While the taxi data is provided as a list of events (one row per taxi ride), the weather data might be on the granularity of minutes, hours, or days, and might also contain missing information. 
To join the taxi dataset with the NYC weather dataset, one needs to define an appropriate join key and potentially pre-aggregate and interpolate the weather dataset, as well as deal with null values. 
Moreover, there might be hundreds of related tables, creating a large amount of work for the user.

There exists a unique opportunity to explore this type of data augmentation automatically as part of an automatic machine learning pipeline.
Automatically finding and joining related tables is challenging~\cite{fernandez2018aurum}. 
Users often have a hard time distinguishing a semantically meaningful join from an meaningless join if they don't know details about the join itself; the relevance of a join  depends on the semantics of the table and there is no easy way to quantify semantics. 
However, in the context of automated machine learning, the automating joins is significantly simpler as a clear evaluation metric exists: does the prediction performance (e.g., accuracy, F1 score, etc.) after the join increase? This sidesteps the tricky question of the semantic connection between two tables (or perhaps implicitly measures it) while still increasing the metric the user cares most about, the predictive model's performance.

In this paper, we explore how we can extend existing automatic machine learning tools to also perform automatic data augmentation by joining related tables. 
A naive solution to the problem would try to find common join keys between tables, then join all possible tables to create one giant ``uber''-table, and then use a feature selection method (which are already part of many \AML-tools) to select the right features. 
Obviously, such an approach has several problems. 
First, in some cases like the weather and taxi data, a direct join-key might not exist, requiring a more fuzzy type of join. 
Second, joining everything might create a giant table, potentially with more features than rows. This can be problematic even for the best existing feature selection methods (i.e., it creates an underdetermined system). Even with ideal feature selectors, the size of such an ``uber''-table may significantly slow down the \AML process.

Thus, we developed \system{}, a first Automatic Relational Data Augmentation system. From a base table provided by a user and a foreign key mapping (provided by the user or discovered with an external tool~\cite{fernandez2018aurum}), \system{} automatically and efficiently discovers joins can help model performance, and avoids joins that add little or no value (e.g., noise). At a high level, \system{} works by first constructing a \emph{coreset},
a representative but small set of rows from the original sample table. This coreset is strategically joined with candidate tables, and features are evaluated using a novel technique we call \emph{random injection feature selection (\textsf{RIFS})}. At a high level, \textsf{RIFS} helps determine if the results of a join are helpful for a predictive model by comparing candidate features against noise: if a candidate feature performs no better than a feature consisting of random noise, that feature is unlikely to be useful. \system{} handles joins between mismatched keys that may not perfectly align (e.g., time) using several interpolation strategies. The final output of \system{} is an augmented dataset, containing all of the user's original dataset as well as additional features that increase the performance of a predictive model.

While there has been prior work on data mining, knowledge discovery, data augmentation and feature selection, to the best of our knowledge no system exists which automatically explores related tables to improve the model performance. 
Arguably most related to our system are Kumar et. al.~\cite{kumar2016join} and Shah et al.~\cite{shah2017key}.
Given a set of joins, the authors try to eliminate many-to-one (n-to-1) joins which are not necessary (i.e., highly unlikely to improve a predictive model) because the foreign key already contains all the information from the external table. 
For example, consider again the taxi dataset, but where the start address is modeled as a FK-PK relationship to a street-table. 
In that case, depending on the model, the join to the street-table might not be necessary as the foreign-key acts as an embedding for the street. In other words, the foreign key (street address) already contains the street name, which is the only extra piece of information that could be gained from joining with the street-table. 

However, the techniques in \cite{kumar2016join,shah2017key} focus only on classification tasks and err on the side of caution, only eliminating joins that are statistically highly unlikely to improve the final model. In other words, these techniques focus on excluding a narrow subset of tables that will not improve a model: just because a table is not ruled out by the techniques of~\cite{kumar2016join,shah2017key} does not mean that this table will improve model performance. In fact, the techniques of~\cite{kumar2016join,shah2017key} are intentionally conservative.  \system is designed to automatically find feature sets that actually improve the predictive power of the final model.
Furthermore, they can only eliminate many-to-one joins and neither deal with
one-to-many joins, nor with fuzzy joins (e.g., the weather and taxi data which must be joined on time). 
Thus, both \cite{kumar2016join} and~\cite{shah2017key} are orthogonal to this work, as we are concerned with effectively augmenting a base table to improve model accuracy as opposed to determining which joins are potentially safe to avoid.
In our experimental study, we demonstrate that Kumar et al.'s decision rules can be used as a prefiltering technique for \system{}, slightly affecting model performance (sometimes better, sometimes worse) but always improving runtime.

In summary we make the following contributions:
\begin{itemize}[leftmargin=*]
\item{We introduce \system{}, a system for automatic relational data augmentation which can discover joins that improve the performance of predictive models,}
\item{we propose simple methods for performing one-to-many and ``soft key'' (e.g., time) joins useful for ML models,}
\item{we introduce \textsf{RIFS}, a specific feature selection technique custom-tailored to relational data augmentation,}
\item{we experimentally demonstrate that effectiveness of our prototype implementation of \system{}, showing that relational data augmentation can be performed automatically in an end-to-end fashion.} 
\end{itemize}



%% file: system_model.tex
\section{Problem Overview}
\label{sec:system_overview}
In this section, we describe the setup and problem statement we consider. Our system is given as input a \emph{base table} with labelled data (that is, a specified column contains the target labels) and the goal is to perform inference on this dataset (i.e., be able to predict the target label for a new row of the table). We assume that the base table provided by the user is one table in a (potentially large) data repository (e.g.~\cite{url-auctus}). We assume the learning model is also specified and is computationally expensive to train. For example, random forest models with a large number of trees or SVM models over large datasets can be time consuming to train. \system is otherwise agnostic to the model training process, and can use any type of model training, including simple models like random forest as well as entire \AML systems. 


Our goal then is to \emph{augment} our base table---to add features such that we can obtain a non-trivial improvement on the prediction task. To this end, we assume that an external \emph{data discovery system} (e.g.~\cite{fernandez2018aurum}) automatically determines a collection of \emph{candidate joins}: columns in the base table that are potentially foreign keys into another table. Since data discovery systems like~\cite{fernandez2018aurum} generally depend on a ``human-in-the-loop'' to eliminate false positives, \system is designed to handle a generated collection that is potentially very large and highly noisy (i.e., that the majority of the joins are semantically meaningless and will not improve a predictive model).

\system considers candidate joins based on two types of candidate foreign keys: \emph{hard keys}, which are foreign keys in the traditional sense of a database system and can be joined with their corresponding table using traditional algorithms, and \emph{soft keys}, foreign keys that may not precisely match the values in the corresponding table. For example, a data discovery system may indicate that a column representing time in the user's base table is a potential foreign key into another table storing weather data. If the user's base table has timestamps at a one-day level of granularity, but the weather table has timestamps at a one-minute level of granularity, joining the two tables together requires special care. Simply using a traditional join algorithm may result in many missing values (when granularity does not precisely align), or semantically meaningless results (e.g., if one arbitrarily determines which weather entry to match with each row of the user's base table). To support joins of such soft keys, \system provides the user with several options, including nearest neighbor, linear interpolation, and resampling techniques. We assume that the ``hardness'' or ``softness'' of a join key is indicated by the data discovery system~\cite{fernandez2018aurum}, and we discuss the details of soft key joins in Section~\ref{sec:joins}.

\begin{figure*}
  \centering
  \includegraphics[width=0.80\textwidth]{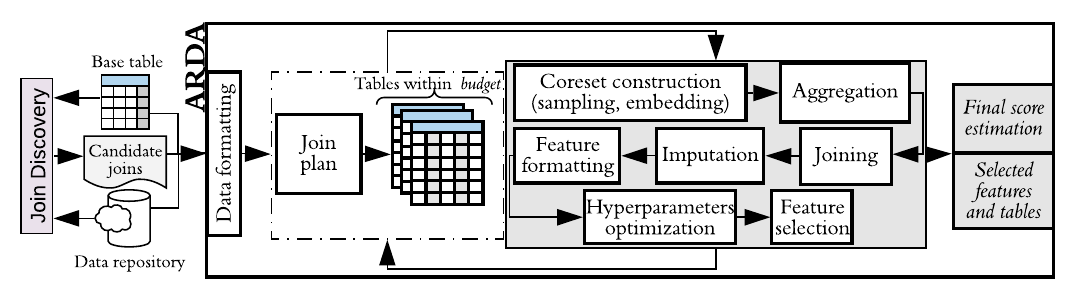}
  \caption{We provide a visual description of the workflow of our system. We start with a database and a base table as input to a join discovery framework that returns a large collection of tables which may contain information relevant to our learning task. After appropriate pre-processing, \system{} joins the candidate tables in batches. }
  \label{fig:system_overview}
\end{figure*}

\begin{figure}[htb!]
  \centering
  \includegraphics[width=0.40\textwidth]{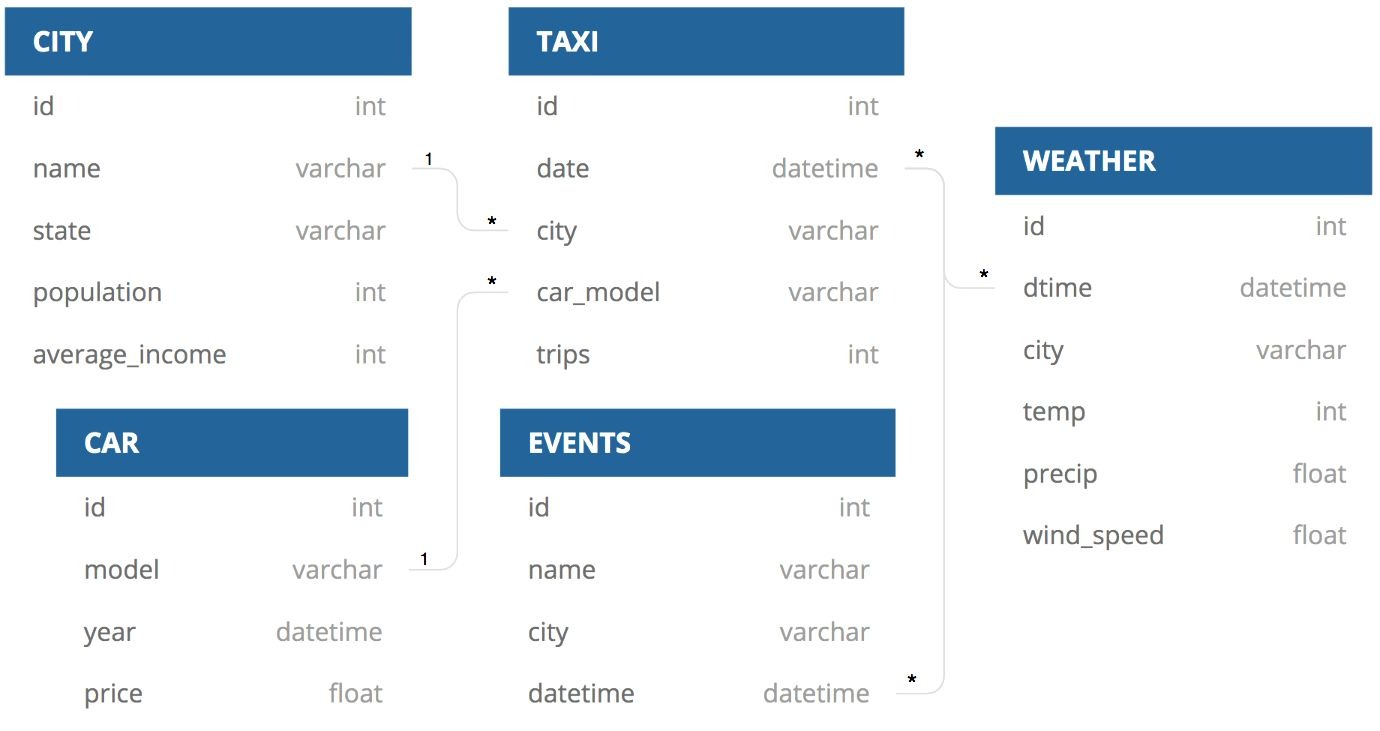}
  \caption{Example Schema: Initially a user has a base table \textsf{TAXI} and she finds a pool of joinable tables to see if some of them can help improve prediction error for taxi demand for a specific date given in target column \textsf{trips}.}
  \label{fig:example_schema}
\end{figure}

\textbf{Example.} 
Traditionally, a user has had to invest significant effort in deciding what data can usefully augment a classification problem.  For example, suppose that
given a base table \textsf{TAXI}, the learning task is to predict the target column \textsf{trips} that gives the number of trips a given a taxi made or will make for a given day. Traditionally, a user might speculate that the weather might significantly impact demand. Since weather data is available in the user's data repository, she can writing code to join the \textsf{TAXI} and \textsf{WEATHER} datasets on columns \textsf{date} and \textsf{time}. After determining a strategy to join together these soft keys, she then evaluates whether or not the additional weather data improved model accuracy. Afterwards, she can continue to search for other potential augmentations, such as sporting event schedules or traffic data. 

The goal of this project is to automate this arduous process with little to no user intervention from the user.  In Figure~\ref{fig:example_schema} we show an example schema representing foreign keys discovered by a data discovery system on a (potentially heterogeneous) data repository. In reality, this schema would be much bigger, since it would contain a large quantity of irrelevant datasets. Our task is to discover tables that contain information that can improve a predictive model's performance, without requiring a human being to decide which candidate joins are potentially relevant. The central questions that arise here include how to  effectively combine information from multiple relations, how to efficiently determine which other relations provide valuable information and how to process a massive number of tables efficiently. 

Once the relevant join is performed, we must contend with the massive number of irrelevant features interspersed with a small number of relevant features. We  observed that after many joins,
the resulting dataset often results in predictive models with worse performance than training a model with no augmentation. This is because machine learning models are can be misled by noise, and large numbers of features provide more opportunity for such confusion. Therefore, our implementation must efficiently select relevant features in the presence of many irrelevant or noisy features. The rest of this work is devoted to describing our system and how it addresses the aforementioned challenges. 

%% file: joins.tex
\section{Augmentation workflow}
\label{sec:augmentation_workflow}
We begin with a high-level description of the workflow that our system follows. 

\sparagraph{Input to \system{}.}
\system{} requires a reference to a database and a collection of candidate joins from a data discovery system:
a description of the columns in the base table that can be used as foreign keys into other tables.
Often, data discovery systems (e.g., ~\cite{fernandez2018aurum, url-auctus}) provide a ranking of the candidate joins based on expected relevancy (usually determined by simple heuristics).
While such an ordering may not directly correspond to an importance of given a table to downstream learning tasks, \system can optionally make use of this information to prioritize its search. 

\sparagraph{Coreset construction.} As a first part of our joining pipeline we proceed to create a representative subset. We construct a \emph{coreset}, a representative sample of rows from the base table. This sampling is done for computational efficiency: if the base table is small enough, no sampling is required. \system{} allows several types of coreset construction, described in Section~\ref{subsection:coreset}. If sample size is specified, \system{} samples rows according to a customizable procedure, the default being uniform sampling. \system{} allows user to specify the desired coreset size, or \system can automatically select a coreset size using simple heuristics.

\sparagraph{Join plan.} During the stage of a \emph{join plan} \system{} decides in which order tables should be considered for augmentation, how many tables to join at a time, and what tables should be grouped together for further feature selection. \system{} has three available options for a join plan described in Section \ref{sec:joins}. By default, \system uses the \emph{budget} strategy. 

\sparagraph{Join execution.} After determining a join plan, \system begins testing joins for viable augmentations. Performing each join requires special care for (1) handling soft keys, (2) handling missing values, and (3) handling one-to-many joins which might duplicate certain training examples, biasing the final model. We discuss join execution in Section~\ref{sec:joins}.

\sparagraph{Feature Selection.} 
When considering whether or not a particular join represents a useful augmentation, \system{} uses a feature selection process. 
\system{} considers  various types of feature selection algorithms that can be run simultaneously, which we discuss in subsequent sections. These methods include convex and non-linear models, such as Sparse Regression and Random Forests.
Unless explicitly specified, \system{} uses a new feature selection algorithm method that we introduce in Section~\ref{sec:random_injection_feature_selection}. This algorithm, random injection feature selection (\textsf{RIFS}), is based on injecting random noise into the dataset. We compare the running time and accuracy for different methods Section \ref{sec:experiments}.

\sparagraph{Final estimate.} 
Finally, after \system has processed the entire join plan and selected a set of candidate joins to use as augmentations, \system trains a machine learning model on the newly collected features. \system is agnostic to the ML training process, and in our experimental analysis we tested both random forest models and advanced \AML systems~\cite{10.1145/3299869.3319863, url-msazure} AutoML system as an optional estimator.

\subsection{Coreset Constructions}
\label{subsection:coreset}
In this subsection, we discuss various approaches used by \system{} to sample rows of our input to reduce the time we spend on joining, feature selection, model training, and score estimation. While we discuss generic approaches to sampling rows, often the input data is structured and the downstream tasks are known to the users. In such a setting, the user can use specialized coreset constructions to sample rows. We refer the reader to an overview of coreset constructions in \cite{phillips2016coresets} and the references therein.

Coreset construction can be viewed as a technique to replace large data set with a smaller number of representative points. This corresponds to reducing the number of rows (training points) of our input data, which in turn allows us to run feature selection more quickly, possibly at a cost in accuracy.  We consider two main techniques for coreset construction: sampling and sketching. Sampling, as the name suggests, selects a subset of the rows and re-weights them to obtain a coreset. On the other hand,
Sketching relies on taking sparse linear combination of rows,
which inherently results in modified row values. This limits our use of sketching before we join tables since the sketched data may result in joins that are vastly inconsistent with the original data. 

\sparagraph{Uniform Sampling.} The simplest strategy to construct a coreset is uniformly sampling the rows of our input tables. This process is extremely efficient since it does not require reading the input to create a coreset. 
However, uniform sampling does not have any provable guarantees and is not sensitive to the data. It is also agnostic to outliers, labels and anomalies in the input.  For instance, if our input tables are labelled data for classification tasks, and one label appears way more often than others, a uniform sample might completely miss sampling rows corresponding to certain labels. Thus, the sample we obtain may not be diverse or well-balanced, which can negatively impact learning. 

\sparagraph{Stratified Sampling.} To address the shortcomings of uniform sampling, we consider stratified sampling.  Stratification is the process of dividing the input into homogeneous subgroups before sampling, such that the subgroups form a partition of the input. Then simple random sampling or systematic sampling can be applied within each stratum. 

The objective is to improve the precision of the sample by reducing sampling error. It can produce a weighted mean that has less variability than the arithmetic mean of a simple random sample of the population. For classification tasks, if we stratify based on labels and use uniform sampling within each stratum, we obtain a diverse, well-balanced sub-sample, and no label is overlooked. 

\sparagraph{Matrix Sketching.} Finally, consider sketching algorithms to sub-sample the rows of our input tables. Sketching has become a useful algorithmic primitive in many big-data tasks and we refer the reader to a recent survey \cite{woodruff2014sketching}, though is only applicable to numerical data. We note that under an appropriate setting of parameters, sketching the rows of the input data approximately preserves the subspace spanned by the columns. 

An important primitive in the sketch-and-solve paradigm is a subspace embedding \cite{clarkson2013low,nelson2013osnap}, where the goal is to construct a concise describe of data matrix that preserves the norms of vectors restricted to a small subspace. Constructing subspace embeddings has the useful consequence that accurate solutions to the sketched problem are approximate accurately solutions to the original problem. 

\vspace{-0.1in}
\begin{definition}(Oblivious subsapce embedding.)
\label{def:subspace_embedding}
Given $\epsilon, \delta>0$ and a matrix $A$, a distribution $\mathcal{D}(\epsilon, \delta)$ over $\ell \times n$ matrices $\Pi$ is an oblivious subspace embedding for the column space of $A$ if with probability at least $1-\delta$, for all $ x \in \mathbb{R}^d $, 
$(1-\epsilon)\|A x\|_2 \leq \| \Pi A x \|_2 \leq (1+\epsilon)\|Ax\|_2 $.
\end{definition}
We use OSNAP matrix for $\Pi$ from \cite{nelson2013osnap}, where each column has only one non-zero entry. In this case, $\Pi A$ can be computed in $nnz(A)\log(n)$ time, where nnz denotes the sparsity (number of nonzero entries) of $A$. 
\vspace{-0.1in}
\begin{definition}(OSNAP Matrix.)
Let $\Pi \in \mathbb{R}^{\ell \times n}$ be a sparse matrix such that for all $i \in [n]$, we pick $j \in [\ell]$ uniformly at random, such that $\Pi_{i,j} = \pm 1$ uniformly at random and repeat $\log(n)$ times.
\end{definition}

For obtaining a subspace embedding for $A$, $\Pi$ needs to have $\ell=d\log(n)/\epsilon^2$ rows. We note that for tables where the number of samples (rows) is much larger than the number of features(columns) the above algorithm can be used to get an accurate representation of the feature space in nearly linear time. However, here we note that $\Pi$ takes linear combinations of rows of $A$ and thus does not preserve numeric values. While the above guarantees hold in the worst-case, for real-world data, we can often use a lot fewer rows in our sketch. 

Since sketching methods apply linear combinations of rows, we cannot hope to sketch the base table before the join takes place, without modifying it's contents. 
Therefore, \system{} sketches tables after the join is performed. 
Note that \system{} binarizes categorical features into a set of numerical features, which are amenable to sketching. For classification tasks, \system{} sketch rows independently within each label, analogous to stratified sampling.
\vspace{-0.1in}
\section{Joins}
\label{sec:joins}

After selecting a coreset, \system{} needs to determine which tables will produce valuable augmentations. \system{} provides different strategies for performing joins, as discussed below.
The objective of the join is to incorporate information from a foreign table in the optimal way. A key requirement from our join procedure is to preserve all base table rows since we do not want to artificially add or remove training examples.  

\sparagraph{Joins.}
There are four common join types between two tables \textsf{INNER JOIN}, \textsf{FULL JOIN}, \textsf{RIGHT JOIN}, and \textsf{LEFT JOIN}.  But for our application, the goal is to \emph{add} information to \emph{each} row of the base table.  Losing base table rows would lose data, and creating rows that do not correspond to base table rows would not be meaningful.  Thus, only certain joins are suitable.

\textsf{LEFT JOIN} selects all records from the left table (base table), and the matched records from the right table (foreign table). The result is NULL from the right side if there is no match. This is the only join type that works for our augmentation task since it both preserves every record of the base table and brings only records from the foreign table that match join key values in base table. \textsf{LEFT JOIN} semantics also include leaving NULL when there is no match and results in missing values in the resulting dataset. There are many standard data imputation techniques in ML that we can use to handle these missing values.

None of the other join types satisfy our requirements. Consider for example the \textsf{INNER JOIN} which selects only records that have matching values in both tables. This type of join will drop training examples from the base table that do not join---losing data. For example, assume relation \textsf{TAXI} from Figure \ref{fig:example_schema}  
performed an inner join with relation \textsf{CAR} on \textsf{car model} attribute. Further assume there is only a single car model in \textsf{TAXI} that is found in \textsf{CAR}. In this scenario, all the rows with car models that do not exist in \textsf{CAR} would be lost.

\sparagraph{Key Matches.}
 \system{} supports single key join, multiple key join (composite keys), mixed key join (composite key consists of soft and hard keys), and multiple-option key join (table has different keys it can be joined on with base table). The latter case implies different joining options with the foreign table. In this scenario \system{} joins on each key separately. An alternative option is to join on thr key that gives a larger intersection, but this strategy could fail if such join happens to be useless for a target learning task.

In \system{}, we handle joining on \emph{hard keys}, \emph{soft keys}, and a custom combinations of them. 
Joining on a \emph{hard key}, like Restaurant ID, implies joining rows of two tables where the values in column-keys are an exact match. 
When we join on a \emph{soft key} we do not require an exact match between keys. Instead, we join rows with the column corresponding to the closest value. 
Examples of \emph{soft keys} include time, GPS location, cities, age etc.  


We note that in the special case of \system{} receiving keys from the time series data it automatically performs soft join. \system{} has two settings for soft join:

\textit{Nearest Neighbour} Join: This joins a base table row value with the nearest value in the foreign table. The distance to rows the foreign table is defined in terms of numerical quantities based on the \emph{soft key}. If tolerance threshold is specified and nearest neighbour does not satisfy the threshold then \emph{null} values are filled instead.

\textit{Two-way nearest neighbour}
Join: For a given value of (the join column) of a base table row, this method finds the row largest foreign key less than the value and the row with smallest key greater than the value.  These two rows from the foreign table are combined into one using linear interpolation on numeric values. For instance, let $x$ be the numerical value for the base table key and $y_{\textrm{low}}, y_{\textrm{high}}$ be the values of the foreign keys that matched corresponding to rows $r_{\textrm{low}}$ and  $r_{\textrm{high}}$. Then, 
\[
x = \lambda y_{\textrm{low}} + (1-\lambda)y_{\textrm{high}}
\]
for some $\lambda \in [0,1]$. We then join the row in the base table with $\lambda r_{\textrm{low}} + (1-\lambda) r_{\textrm{high}}$.
This is used to account for how distant the keys are from the base table key value. If the values are categorical, they are selected uniformly at random.

\sparagraph{Time-Resampling.} Consider a situation when the base table has time series data specified in \textsf{month/day/year} format, while foreign table has format \textsf{month/day/year hr:min:sec}. In order to perform a join the time data needs to be ``aligned'' between the two tables.

One option would be to resolve the base table format to the midnight timestamp: \textsf{month/day/year 00:00:00}. However, for a hard join this might result in no matches with a foreign table and for a soft Nearest Neighbour join this would result in joining with only one row from a foreign table that has closest time to a midnight for the same day. This situation would result in information loss 
and therefore affect the quality of resulting features. \system{} identifies differences in time granularity and aggregates data over the span of time of a less precise key.
In our scenario all rows that correspond to the same day would be resampled (aggregated) 
in foreign table before the join takes place.

\sparagraph{Table grouping.} \system{} supports grouping tables at three levels of granularity for joining:
\begin{enumerate}[leftmargin=*]
    \item \emph{Table-join}: One table at a time in the priority order specified by the input. Based on our experiments it is the least desirable type of join since it adds significant time overhead and does not capture co-predicting features partitioned across tables.
    \item \emph{Budget-join:} As many tables at a time as we can fit within a predefined  \emph{budget}. The budget is a user defined parameter, set to be the number of rows by default. Size of \emph{budget} trades off the number of co-predicting features we might discover versus the amount of noise the model can tolerate to distinguish between good and bad features.
    \item \emph{Full materialization join:} All the tables prior to performing feature selection.
\end{enumerate}

Table grouping is used by \system{} to create a \emph{join plan} that is iteratively executed until all tables are processed or user-specified accuracy/error is achieved. 
By default, \system{} uses provided scores by Join Discovery system,  such as Aurum \cite{fernandez2018aurum} (or NYU Auctus), or if missing, \system{} computes \emph{intersection-score}.  The \emph{Tuple Ratio} from \cite{kumar2016join} scoring can be specified by the user on demand. Tables are grouped in batches such that one batch does not exceed allowed 
\emph{budget}.
In \system{} \emph{budget} is defines as maximum number of features we process at a time. By default, \emph{budget} equals coreset size. An exception to this rule happens when a single table has more features than rows in a coreset, in this case \system{} ships an entire table to a feature selection 
pipeline.

\sparagraph{Join Cardinality.}
There are four types of join cardinality: \textit{one-to-one}, \textit{one-to-many}, \textit{many-to-one}, and \textit{many-to-many}. Both \textit{one-to-one} and \textit{many-to-one} preserve the initial distribution of the base table (training examples) by avoiding changes in number of rows. Recall, a  \textit{one-to-one} join matches exactly one key in base table to exactly one key in a foreign table. Similarly, a \textit{many-to-one} join matches many keys in our base table to one row from foreign table, which again suffices.

However, for \textit{one-to-many} and \textit{many-to-many} joins, we would need to match at least one record from base table to multiple records from foreign table. This would require repeating the base table records to accommodate all the matches from a foreign table and introduce redundancy. Since we cannot change the base table distribution, such a join is infeasible.
To address the issue with \textit{one-to-many} and \textit{many-to-many} joins we pre-aggregate foreign tables on join keys, thereby effectively reducing to the \textit{one-to-one} and \textit{many-to-one} cases.

\sparagraph{Imputation.}
Missing data is a major challenge for modern Big-Data systems. This data can be  real, Boolean, or ordinal. Canonical examples of missing data appears in Netflix recommendation system data, health surveys, census questionnaires etc. Common imputation techniques include hot deck, cold deck and mean substitution. The deck methods are heuristics, where missing data is replaced by a randomly selected similar data point. More principled methods assume the data is generated from a simple parametric statistical model. The goal is to then learn the the best fit parameters for this model.

Since in data augmentation workflow we do not assume anything about the input data, we work with simple approaches that reduce the total running time of the system. We implemented a simple imputation technique: we use the median value for numeric data and uniform random sampling is used for categorical values.

%% file: techniques.tex
\section{Feature Selection Overview}
\label{sec:feature_selection_overview}

Feature selection algorithms can be broadly categorized into filter models, wrapper models 
and embedded models. The filter model separates feature selection from classifier learning 
so that the bias of a learning algorithm does not interact with the bias of a feature 
selection 
algorithm. Typically, these algorithms rely on 
general characteristics of the training data such as distance, consistency, dependency, 
information, and correlation. Examples of such methods include Pearson correlation coefficient, Chi-squared test, mutual information and numerous other statistical significance tests~\cite{feature_selection}. We note that filter methods only look at the input features and do not use any information about the labels, and are thus sensitive to noise and corruption in the features.

The wrapper model uses the predictive accuracy of the learning algorithm to determine the 
quality of selected features. Wrapper-type feature selection methods are tightly
coupled with a specific classifier, such as correlation-based feature selection (CFS) \cite{guyon2002gene} and support vector machine recursive feature elimination (SVM-RFE) \cite{hall1999correlation}. 
The trained classifier is then used select a subset of features. Popular approaches to this include \textit{Forward  Selection}, \textit{Backward Elimination} and \textit{Recursive Feature Elimination (RFE)}, which
iteratively add or remove features and compute performance of these subsets on learning tasks. Such methods are likely to get stuck in local minimax  
\cite{john1994irrelevant,yang1998feature,hall1999correlation}.  Further, forward selection may ignore weakly correlated 
features and backward elimination may erroneously remove relevant features due to noise.  They often have good performance, but their computational cost is very expensive for large data and training massive non-linear models \cite{liu2010feature,estevez2009normalized}.

Given the shortcomings in the two models above, we focus on the embedded model, which includes information about labels, and incorporates the performance of the model on holdout data. Typically, the first step is to obtain a ranking of the features by optimizing a convex loss function \cite{cawley2007sparse,liu2005toward}. Popular objective functions  include quadratic loss, hinge loss and logistic loss, combined with various regularizers, such as $\ell_1$ and elastic net. The resulting solution is used to select a subset of features and evaluate their performance. This information is then used to pick a better subset of features.

One popular embedded feature selection algorithm is Relief, which is considered to be effective and efficient in practice \cite{kira1992practical,robnik2003theoretical,sun2007iterative}. 
However, a crucial drawback of Relief is that in the presence of noise, performance degrades severely \cite{sun2007iterative}. Since Relief relies on finding nearest-neighbors in the original feature space, having noisy features can change the objective function and converge to a solution arbitrarily far from the optimal. 
While \cite{sun2007iterative}  offers an iterative algorithm to fix Relief, this requires running Expectation-Maximization (EM) at each step, which has no convergence guarantees. Further, each step of EM takes time quadratic in the number of data points. Unfortunately, this algorithm quickly becomes computationally infeasible on real-world data sets. 

Given that all existing feature selection algorithms that can tolerate noise are either computationally expensive, use prohibitively large space or both, it's not obvious if such methods could be effective in data augmentation scenario when we deal with massive number of spurious features. 

\vspace{-0.15in}
\section{Random Injection Based\\
Feature Selection}
\label{sec:random_injection_feature_selection}

In this section, we describe our random injection based feature selection algorithm, including the key algorithmic ideas we introduce.  Recall, we are given a dataset, where the number of features are significantly larger than the number of samples and most of the features are spurious. Our main task is to find a subset of features that contain signal relevant to our downstream learning task and prune out irrelevant features. 



This is a challenging task since bereft of assumptions on the input, any \textit{filter based} feature selection algorithm would not work since it does not take prediction error of a subset of features into account. For instance, consider a set of spurious input features that have high Pearson Correlation Coefficient or Chi-Squared test value.
Selecting these features would lead to poor generalization error in our learning task but filter methods do not take this information into account. 
To this end, we design a comparison-based feature selection algorithm that circumvents the requirement of testing each subset of features. We do this by injecting carefully constructed random features into our dataset. We use the random features as a baseline to compare the input features with.

We train an ensemble of random forests and linear model to compute a joint ranking over the input and injected features. Finally, we use a wrapper method on the resulting ranking to determine a subset of input features contain signal. However, our wrapper method only requires training the complex learning model a constant number of times.  We discuss each part in detail below:

\begin{Frame}[\textbf{Algorithm \ref{alg:robust_feature_selection}} :  Feature Selection via Random Injection]
\label{alg:robust_feature_selection}
\textbf{Input}: An $n \times d$ data matrix $A$, a threshold $\tau$ and fraction of random features to inject $\eta$, the number of random experiments performed $k$.  
\begin{enumerate}
    \item Create $t = \eta d$ random feature vectors $n_1, n_2, \ldots n_t$  $\in \mathbb{R}^n$ using Algorithm \ref{alg:noise_injection}. Append the random features to the input and let the resulting matrix be $A' = \left[A \mid N \right]$.
    \item Run a feature selection algorithm (see discussion below) on $A'$ to obtain a ranking vector $r \in [0,1]^{d+t}$.  Repeat $k$ times. 
    \item Aggregate the number of times each feature appears in front of all the random features $n_1, \ldots n_t$ according to the ranking $r$. Normalize each value by $k$. Let $r^{*}\in [0,1]^d$ be the resulting vector. 
\end{enumerate}
\textbf{Output:} A subset $\mathcal{S} \subseteq [d]$ of features such that the corresponding value in $r^*$ is at least $\tau$. 
\end{Frame}

\subsection{Random Feature Injection}
Given that we make no assumptions on the input data, our implementation should capture the following extreme cases: when most of the input features are relevant for the learning task, we should not prune out too many features. On the other hand, when the input features are mostly uncorrelated with the labels and do not contain any signal, we should aggressively prune out these features.  We thus describe a random feature injection strategy that interpolates smoothly between these two corner cases.

We show that in the setting where a majority of the features are good, injecting random features sampled from the standard Normal, Bernoulli, Uniform or Poisson distributions suffices, since our ensemble ranking can easily distinguish between random noise and true features. The precise choice of distribution depends on the input data.  

The challenging setting is where the features constituting the signal is a small fraction of the input. Here, we use a more aggressive strategy with the goal of generating random features that look a lot like our input. To this end, we fit a statistical feature model that matches the empirical moments of the input data. Statistical modelling has been extremely successful in the context of modern machine learning and moment matching is an efficient technique to do so. We refer the reader to the surveys and references therein \cite{mccullagh2002statistical,davison2003statistical,khalid2014survey, sun2019tutorial}.

We compute the empirical mean $\mu = \frac{1}{d}\sum_{i\in[d]}A_{*,i}$ and the empirical covariance $\Sigma = \frac{1}{d}\sum_{i\in[d]}(A_{*,i}-\mu) (A_{*,i}-\mu)^T$ of the input features. 
Intuitively, we assume that the input data was generated by some complicated probabilistic process that cannot be succinctly describes and match the empirical moments of this probabilistic process. An alternative way to think about our model is to consider $\mu $ to be a typical feature vector with $\Sigma$ capturing correlations between the coordinates. We then fit a simple statistical model to our observations, namely $\mathcal{N}(\mu, \Sigma)$. Finally, we inject features drawn i.i.d. from $\mathcal{N}(\mu, \Sigma)$.

\begin{Frame}[\textbf{Algorithm \ref{alg:noise_injection}} : Random Feature Injection Subroutine]
\label{alg:noise_injection}
\textbf{Input:} An $n \times d$ data matrix $A$.
\begin{enumerate}
    \item Compute the empirical mean of the feature vectors by averaging the column vectors of A, i.e. let  $\mu = \frac{1}{d}\sum_{i\in[d]}A_{*,i}$.
    \item Compute the empirical covariance  $\Sigma = \frac{1}{d}\sum_{i\in[d]} $ $(A_{*,i}-\mu) (A_{*,i}-\mu)^T$.
    \item We model the distribution of features as $\mathcal{N}(\mu, \Sigma)$ and generate $\eta d$ i.i.d. samples from this distribution.
\end{enumerate}
\textbf{Output:} A set of $\eta d$ random vectors that match the empirical mean and covariance of the input dataset.
\end{Frame}

\subsection{Ranking Ensembles}
We use a combination of two ranking models, Random Forests and regularized Sparse Regression.
Random Forests are models that have large capacity and can capture non-linear relationships in the input. They also typically work well on real world data and our experiments indicate that the inherent randomness helps identifying signal on average. 
However, since Random Forests have large capacity, as we increase the depth of the trees, the model may suffer from over-fitting. Additionally, Random Forests do not have provable guarantees on running time and accuracy. We use an off-the-shelf implementation of Random Forests and \system{} takes care of tuning the hyper-parameters. 


We use $\ell_{2,1}$-norm minimization as a convex relation of sparsity. The $\ell_{2,1}$-norm of a matrix sums the absolute values of the $\ell_2$-norms of its rows. Intuitively, summing the absolute values can be thought of as a convex relaxation of minimizing sparsity. Such a relaxation appears in Sparse Recovery, Compressed Sensing and Matrix Completion problems \cite{candes2006robust,candes2009exact,candes2010matrix,marques2018review} and references therein. 
The $\ell_{2,1}$-norm minimization objective has been considered before in the context of feature selection \cite{stojnic2010ell_, xiao2012proximal,qian2013robust, ma2016robust,liu2018dimension}. Formally, let $X$ be our input data matrix and $Y$ is our label matrix. We consider the following regularized loss function :
\begin{equation}
\label{eqn:opt}
    \mathcal{L}(W) = \min_{W\in \mathbb{R}^{c \times d} } || W X  - Y||_{2,1} + \gamma ||W^T||_{2,1}
\end{equation}
While this loss function is convex (since it is a linear combination of norms), we note that both terms in the loss function are not smooth functions since $\ell_1$ norms are not differentiable around $0$.  Apriori it is unclear how to minimize this objective without using Ellipsoid methods, which are computationally expensive (large polynomial running time). A long line of work studies efficient algorithms to optimize the above objective and use the state-of-the-art efficient gradient based solver from \cite{qian2013robust}
to optimize the loss function in Equation \ref{eqn:opt}. However, this objective does not capture non-linear relationships among the feature vectors. We show that a combination of the two approaches works well on real world datasets. 

Additionally, for classification tasks we consider setting where the labels of our dataset may also be corrupted.
Here, we use a modified objective from \cite{qian2013robust}, where the labels are included as variables in the loss function. The modified loss function fits a consistent labelling that minimizes the $\ell_{2,1}$-norm from Equation \ref{eqn:opt}. We observe that on certain datasets, the modified loss function discovers better features.

\subsection{Aggregate Ranking}

We use a straightforward re-weighting strategy to combine the rankings obtained from Random Forests (RF) and Sparse Regression (SR). Given the aforementionned rankings, we compute an aggregate ranking parameterized by $\nu \in [0,1]$ such that we scale the RF rankings by $\nu$ and the SR ranking by $(1-\nu)$. Given an aggregate ranking, one natural way to do feature selection is \emph{exponential search}. We choose the final number of features using a modified exponential search from \cite{bentley1976almost}: we start with 2 features, and repeatedly double the number of features we test until model accuracy decreases. 
Suppose the model accuracy first decreases when we test $2^k$ features. Then,  we perform a binary search between $2^{k-1}$ and $2^k$. We experimentally observe that even the aggregate rankings are not monotone in prediction error. If instead, we perform a linear search over the ranking, we end up training our model $n$ times, which may be prohibitively expensive (this strategy is also known as forward selection).

\begin{Frame}[\textbf{Algorithm \ref{alg:ranking_subroutine}} : Wrapper Algorithm]
\label{alg:ranking_subroutine}
\textbf{Input:} An $n \times d$ data matrix $A$, a set of thresholds $\Tau$
For each $\tau \in \Tau$ in increasing order:
    \begin{enumerate}
        \item For each group run Algorithm \ref{alg:robust_feature_selection} with $A$ and $\tau$ as input. Let $\mathcal{S}\subseteq [d]$ denote the indices of the features selected by Algorithm \ref{alg:robust_feature_selection}.  
        \item Train the learning model on $A_{\mathcal{S}}$, i.e. the subset of features indexed by $\mathcal{S}$ and test accuracy on the holdout set.
        \item If the accuracy is monotone, proceed to the new threshold. Else, output the previous subset. 
\end{enumerate}
\textbf{Output:} A subset of features indexed by set $\mathcal{S}$.
\end{Frame}

We therefore inject random features into our dataset and compute an aggregate ranking on the joint input. We repeat this experiment $t$ times, using fresh random features in each iteration. We then compute how often each input feature appears in front of all the injected random features in the ordering obtained by the aggregate ranking. Intuitively, we expect that features that contain signal are ranked ahead of the random features on average. Further, input features that are consistently ranked worse than the injected random features are either very weakly correlated or systematic noise.

%% file: experiments.tex
\vspace{-0.1in}
\section{Experiments}
\label{sec:experiments}

In this section, we describe our extensive experimental evaluation. 
We begin by presenting our main experiment that shows achieved augmentation on every real world dataset. Next, we evaluate each component of our system, including various coreset constructions, joining algorithms, feature selection algorithms and quality of data augmentation on real-world datasets. Our final experiment shows how well different feature selectors filter out synthetic noise on micro benchmarks. For evaluation of our experiments we used lightly auto-optimized Random Forest model for classification and regression tasks along with SVM with RBF kernel for classification only tasks, such that the best score achieved was reported.

\begin{figure}[htb!]
    \centering
    \vspace{-.15in}
    \includegraphics[width=0.45\textwidth]{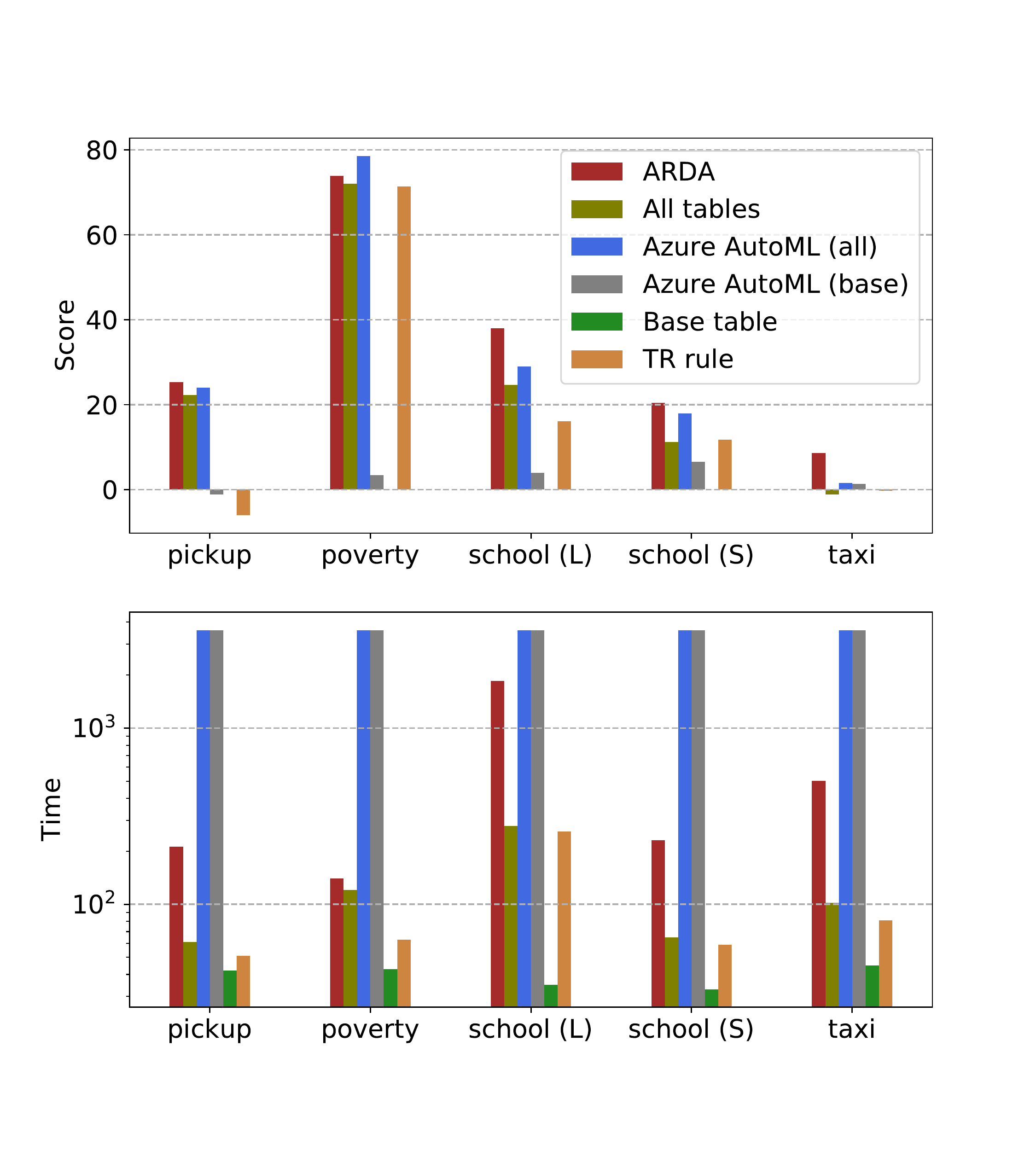}
    \vspace{-.35in}
    \caption{Achieved Augmentation is measured as $\%$ improvement over the baseline accuracy using our default fixed estimator and time is measured in seconds, on a log scale. \textsf{All tables} represents a score achieved using our estimator without feature selection. For \system we used \textsf{RIFS} as a feature selection method. \textsf{TR rule} is a table-filtering method from \cite{kumar2016join}. \textsf{Azure AutoML (all)} represents a score achieved on fully materialized join using Azure AutoML over an hour run.
    \textsf{Azure AutoML (base)} represents a baseline score achieved by Azure AutoML over an hour run.}
    \label{fig:bar_selectors_comparisons_real}
    \vspace{-.1in}
\end{figure}

Recall, our feature selection algorithm consists of a learning model used to obtain a ranking and a subset selection such as forward or backward selection. In our plots, we use the following feature selection methods: \emph{Forward Selection}, \emph{Backward Selection (Backward Elimination)}, \emph{Recursive Feature Elimination (RFE)}, \emph{Random Forest}, \emph{Sparse Regression}, \emph{Mutual Information}, \emph{Logistic Regression}, \emph{Lasso}, \emph{Relief}, \emph{Linear SVM}, \emph{F-test}, \emph{Tuple Rule (TR rule)},  \textsf{RIFS}. We also include experiments with two AutoML systems: Microsoft Azure AutoML \cite{fusi2018probabilistic} and Alpine Meadow \cite{10.1145/3299869.3319863}. 

Methods such as Random Forest, Sparse Regression, Mutual Information, Logistic Regression, Lasso, Relief, and Linear SVM return ranking that we use to select features using repetitive doubling and binary search algorithm. Forward Selection, Backward Selection, and Recursive Feature elimination (RFE) use Random Forest ranker. We picked Random Forest as the main ranker model for a lot of feature selectors because it was showing consistently good performance. 
For comparing ranking algorithms such as Random Forest, Sparse Regression, Mutual Information, Logistic Regression, Lasso, Relief, Linear SVM, F-test, we use \emph{exponential search} described in Section \ref{sec:feature_selection_overview}, as this accurately captures the monotonicity properties of the ranking and performed the best on our data sets.

For our experiments with \textsf{RIFS}, we inject $20\%$ random features ($\mu = 0.2$) drawn i.i.d from Algorithm \ref{alg:noise_injection}. We repeat this process $t$ times (in our experiments $t=10$) and in each iteration we train a Random Forest model as well as a Sparse Regression model to obtain two distinct rankings. For each feature, we compute the fraction of times it appears in front of all random features. We then discard all features less than $\tau$, for $\tau \in \Tau$. 
We only increase the threshold as long as the performance of the resulting features on the holdout set increases monotonically.

\subsection{Real World Datasets}
\label{subsec:real_world_datasets}

Real World datasets are such that given a base table you search open sourced datasets for joinable tables using Join Discovery systems such as Aurum  or NYU Auctus. All of our regression datasets are composed based on base tables provided by the DARPA D3M\footnote{\url{https://www.darpa.mil/program/data-driven-discovery-of-models}} competition. We use the NYU Auctus to search over new tables to augment.  Our composed real scenario datasets:

\begin{figure}[htb!]
    \centering
    \vspace{-.1in}
     \includegraphics[width=0.45\textwidth]{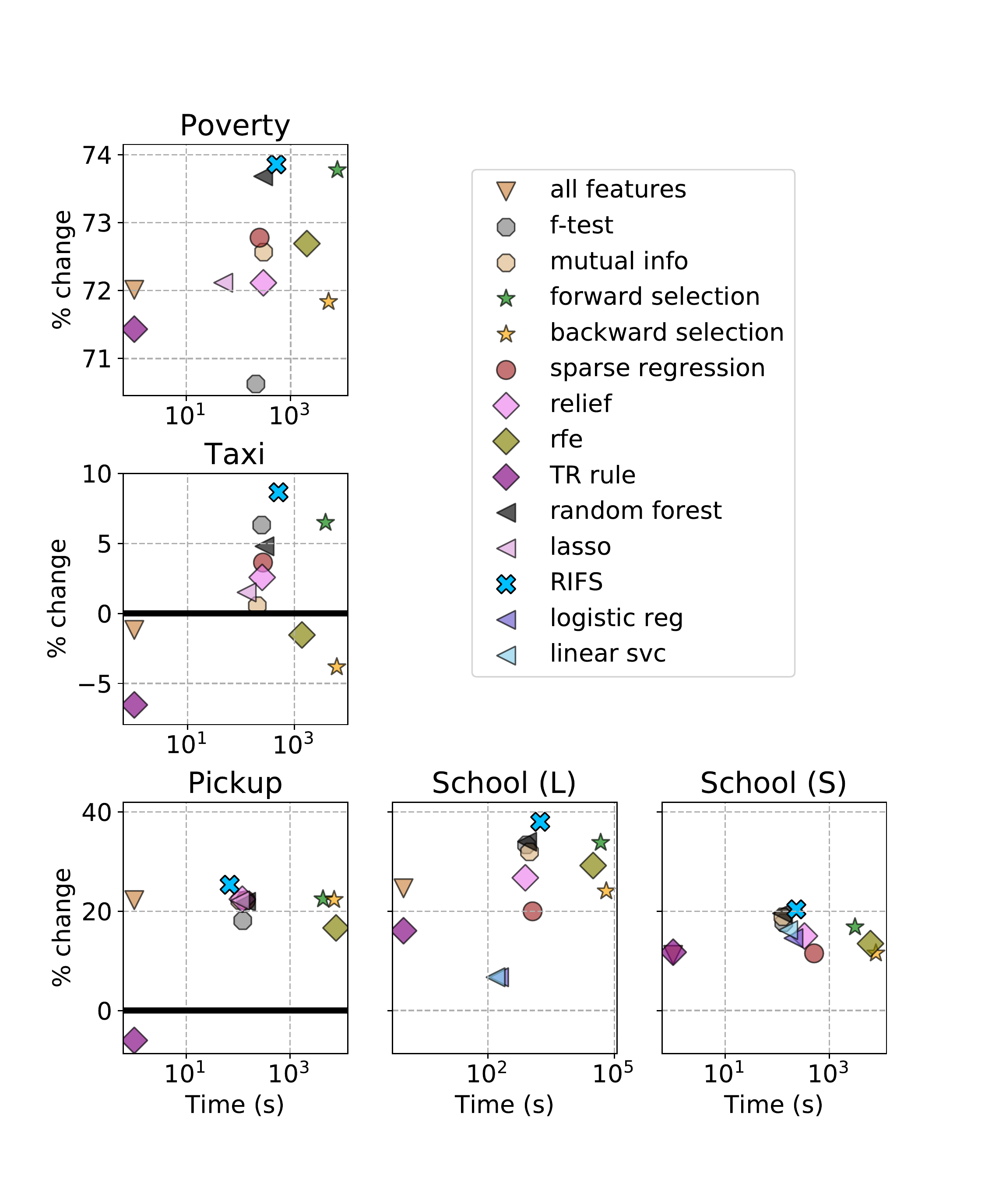}
    \vspace{-.2in}
    \caption{Scores vs. time for real-world datasets, where time along the $x$-axis is in log scale and represents the time spent on feature selection, and score is the $\%$-improvement over the base table accuracy. Since \textsf{all features} and \textsf{TR rule} do not feature selection, the corresponding time is $0$. Our feature selection algorithm, \textsf{RIFS}, consistently performs better augmentation on all datasets.}
    \vspace{-.1in}
\label{fig:scatter}
\end{figure}

\begin{enumerate}[leftmargin=*]
    \item  \textsf{Taxi}, \textsf{Pickup} and \textsf{Poverty}: These are regression datasets contains information about vehicle collision in NYC, taxi demand, and socio-economic indicators  of poverty respectively. The base table is available through NYC Open Data\footnote{\url{https://data.cityofnewyork.us/Public-Safety/NYPD-Motor-Vehicle-Collisions/h9gi-nx95}}, which can be retrieved using the Socrata API\footnote{\url{https://dev.socrata.com/}}. In addition to base table we collected 29 joinable tables. \textsf{Pickup}  contains hourly passenger pickup numbers from LGA airport by Yellow cabs between Jan 2018 to June 2018. The target prediction is the number of passenger pickups for a given hour. In addition to base table we collected 23 joinable tables.  \textsf{Poverty} consists of socio-economic features like poverty rates, population change, unemployment rates, and education levels vary geographically across U.S. States and counties. In addition to base table we collected 39 joinable tables.
    \item \textsf{School (S,L)}: This is a classification dataset from the DataMart API using exact match as the join operation. The target prediction is the performance of each school on a standardized test based on student attributes. In addition to base table, School (S) collected 16 joinable tables whereas School (L) collected 350 joinable tables.
\end{enumerate}


\subsection{Micro Benchmarks}
\label{subsec:micro_benchmarks}

We construct synthetic data to test the performance of feature selection algorithms. These experiments are done since we do not know the ground truth information about the features for a real world datasets that were constructed in a standard augmentation workflow.
Since we do not know ground truth information about the features for a real world datasets we constructed two synthetic datasets to evaluate how well different methods perform in terms of noise filtering. 

\begin{enumerate}[leftmargin=*]
    \item \textsf{Kraken:} A classification dataset consisting of anonymized sensors and usage statistics from the Kraken supercomputer. 
    The target represents machine failure within a $24$-hour window.  The labels are binary, with $568$ samples being $0$ and $432$ being $1$.
    \item \textsf{Digits:} A standard multi-label classification data set with roughly $180$ samples for each digit between $0$ and $9$. It ships with Sklearn\footnote{\url{https://scikit-learn.org/stable/modules/generated/sklearn.datasets.load_digits.html}}.
\end{enumerate}

The synthetic features we append are random, uncorrelated noise vectors sampled from standard distributions such as uniform, Gaussian, and Bernoulli with randomly initialized parameters for these distributions. Since we work in the extreme noise regime, the number of noise features we append is $10\times$ more than the number of original features. We note that not all base table features are relevant and thus may be filtered as well. 

\begin{table*}[t]
\centering
\small
\begin{tabular}{|c|c|c|c|c|c|c||c|c|c|c|}
\hline
\rowcolor[HTML]{EFEFEF} 
\cellcolor[HTML]{EFEFEF} & \multicolumn{2}{c|}{\cellcolor[HTML]{EFEFEF}\textbf{Taxi}} & \multicolumn{2}{c|}{\cellcolor[HTML]{EFEFEF}\textbf{Pickup}} & \multicolumn{2}{c||}{\cellcolor[HTML]{EFEFEF}\textbf{Poverty}} & \multicolumn{2}{c|}{\cellcolor[HTML]{EFEFEF}\textbf{School (S)}} & \multicolumn{2}{c|}{\cellcolor[HTML]{EFEFEF}\textbf{School (L)}} \\ \cline{2-11} 
\rowcolor[HTML]{EFEFEF} 
\multirow{-2}{*}{\cellcolor[HTML]{EFEFEF}\textbf{Method}} & \textbf{\begin{tabular}[c]{@{}c@{}}error \\ x$10^5$\end{tabular}} & \textbf{time} & \textbf{\begin{tabular}[c]{@{}c@{}}error\\ x$10^4$\end{tabular}} & \textbf{time} & \textbf{\begin{tabular}[c]{@{}c@{}}error \\ x$10^6$\end{tabular}} & \textbf{time} & \textbf{accuracy} & \textbf{time} & \textbf{accuracy} & \textbf{time} \\ \hline
baseline (our) & 5658 & 45 & 2088 & 42 & 8116 & 43 & 69.05\% & 33 & 71.11\% & 35 \\ \hline
all features (our) & 572 & 102 & 1623 & 61 & 2271 & 121 & 76.79\% & 65 & 88.61\% & 278 \\ \hline
\begin{tabular}[c]{@{}c@{}}all features\\ (Alpine Meadow)\end{tabular} & 5874 & 3600 & 1761 & 3600 & 2684 & 3600 & 82.25\% & 3600 & 99.73\% & 3600 \\ \hline
\begin{tabular}[c]{@{}c@{}}baseline \\ (Azure AutoML)\end{tabular} & 5584 & 3600 & 2112 & 3600 & 7841 & 3600 & 81.42\% & 3600 & 91.74\% & 3600 \\ \hline
\begin{tabular}[c]{@{}c@{}}all features \\ (Azure AutoML)\end{tabular} & 5572 & 3600 & 1586 & 3600 & 1742 & 3600 & 73.55\% & 3600 & 73.93\% & 3600 \\ \hline
TR rule & 5676 & 81 & 2213 & 51 & 2318 & 43 & 77.17\% & 259 & 82.54\% & 259 \\ \hline
\textsf{RIFS} & 5168 & 553 & 1559 & 168 & 2121 & 575 & 83.13\% & 258 & 98.15\% & 1761 \\ \hline
backward selection & 5874 & 6277 & 1621 & 7181 & 2286 & 5399 & 77.03\% & 8052 & 88.22\% & 64627 \\ \hline
forward selection & 5291 & 3881 & 1619 & 4412 & 2128 & 7974 & 80.67\% & 3221 & 95.16\% & 47491 \\ \hline
RFE & 5745 & 1438 & 1730 & 7799 & 2216 & 2054 & 78.37\% & 6205 & 91.88\% & 31611 \\ \hline
sparse regression & 5452 & 302 & 1624 & 578 & 2209 & 298 & 77.01\% & 594 & 85.32\% & 1151 \\ \hline
random forest & 5386 & 152 & 1630 & 428 & 2136 & 351 & 82.58\% & 152 & 95.29\% & 856 \\ \hline
f-test & 5301 & 243 & 1710 & 232 & 2384 & 281 & 81.26\% & 169 & 94.85\% & 824 \\ \hline
lasso & 5573 & 187 & 1624 & 322 & 2263 & 104 & \multicolumn{4}{c|}{n/a} \\ \hline
mutual info & 5627 & 242 & 1625 & 286 & 2226 & 389 & 82.10\% & 179 & 93.83\% & 1173 \\ \hline
relief & 5512 & 297 & 1619 & 184 & 2263 & 361 & 79.42\% & 381 & 90.12\% & 828 \\ \hline
linear svc & \multicolumn{5}{c}{n/a} & & 80.25\% & 191 & 75.87\% & 201 \\ \hline
logistic reg & \multicolumn{5}{c}{n/a}& & 79.12\% & 263 & 75.87\% & 243 \\ \hline
\end{tabular}
\caption{Results on real world datasets on multiple feature selectors. Error is given as scaled Mean Absolute Error, Time represents feature selection and evaluation time in seconds. }
\label{tab:main_results}
\end{table*}

We compared performance of \textsf{RIFS} with various feature selection methods described in Section \ref{sec:feature_selection_overview}. Figure \ref{fig:scatter} presents percentage of improvement over accuracy achieved on the base table for a given algorithm on a given dataset. As we can see \textsf{RIFS} outperforms all of it's competitors and also performs well in terms of running time. 

Additionally, as we observe, picking all features for the Taxi dataset can even decrease accuracy below that achieved by the base table. The forward selection algorithm is often a close second behind \textsf{RIFS} in terms of accuracy but is at least an order of magnitude slower. The Random Forest ranker with our noise injection rule also achieves augmentation of all datasets and it marginally faster than \textsf{RIFS}.

\begin{figure}[htb!]
    \hspace*{-0.35cm}
    \centering
    \includegraphics[width=0.45\textwidth]{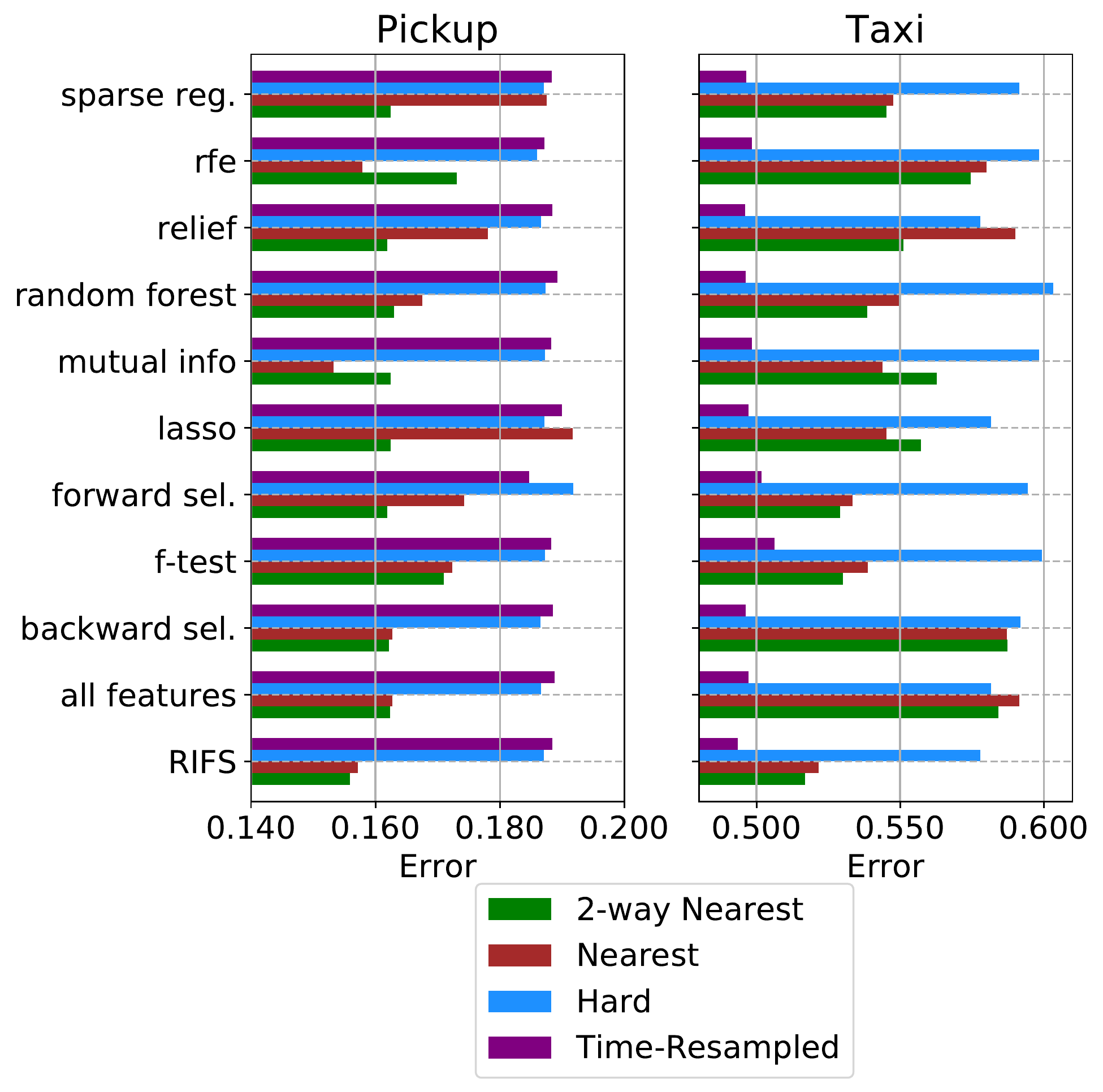}
    \caption{Performance of different techniques for a time series join on multiple feature selectors. \textsf{Pickup} errors were multiplied by $10^2$ and \textsf{Taxi} errors were multiplied by $10$ for convenience.}
    \label{fig:soft_join_types}
    \vspace{-0.1in}
\end{figure}

\begin{figure}[htb!]
    \centering
    \includegraphics[width=0.45\textwidth]{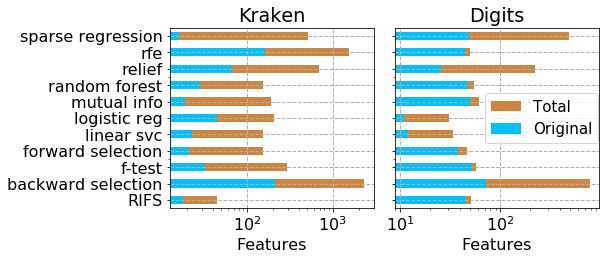}
    \caption{This figure shows the number of features selected by each feature selection method as well as fraction of original features to the total selected features. The difference is planted synthetic noise. X-axis represents the number of features in log scale.}
    \label{fig:bar_selectors_comparisons}
    \vspace{0.1in}
\end{figure}

\subsection{Feature selectors}

\begin{table*}[htb!]
\small
\centering
\begin{tabular}{|c|
>{\columncolor[HTML]{FFCE93}}c |c|
>{\columncolor[HTML]{9AFF99}}c |
>{\columncolor[HTML]{FFCE93}}c |c|
>{\columncolor[HTML]{9AFF99}}c |}
\hline
\cellcolor[HTML]{EFEFEF}                                  & \multicolumn{2}{c|}{\cellcolor[HTML]{EFEFEF}\textbf{School (S)}}                      & \multicolumn{2}{c|}{\cellcolor[HTML]{EFEFEF}\textbf{Digits}}                          & \multicolumn{2}{c|}{\cellcolor[HTML]{EFEFEF}\textbf{Kraken}}                          \\ \cline{2-7} 
\multirow{-2}{*}{\cellcolor[HTML]{EFEFEF}\textbf{Method}} & \cellcolor[HTML]{EFEFEF}\textbf{Stratified} & \cellcolor[HTML]{EFEFEF}\textbf{Sketch} & \cellcolor[HTML]{EFEFEF}\textbf{Stratified} & \cellcolor[HTML]{EFEFEF}\textbf{Sketch} & \cellcolor[HTML]{EFEFEF}\textbf{Stratified} & \cellcolor[HTML]{EFEFEF}\textbf{Sketch} \\ \hline
f-test                                                    & -1.29\%                                     & \cellcolor[HTML]{FFCE93}-1.86\%         & 0.82\%                                      & -3.06\%                                 & \cellcolor[HTML]{9AFF99}0.54\%              & \cellcolor[HTML]{FFCE93}-4.72\%         \\ \hline
mutual info                                               & -3.83\%                                     & \cellcolor[HTML]{FFCE93}-3.35\%         & 0.28\%                                      & -3.69\%                                 & \cellcolor[HTML]{FFCE93}-0.66\%             & \cellcolor[HTML]{FFCE93}-0.56\%         \\ \hline
random forest                                             & -2.84\%                                     & \cellcolor[HTML]{FFCE93}-4.05\%         & 1.05\%                                      & -4.09\%                                 & \cellcolor[HTML]{9AFF99}7.06\%              & 5.84\%                                  \\ \hline
sparse regression                                         & -5.65\%                                     & \cellcolor[HTML]{9AFF99}2.85\%          & \cellcolor[HTML]{FFCE93}-0.15\%                                     & \cellcolor[HTML]{9AFF99}0.04\%         & \cellcolor[HTML]{FFCE93}-2.74\%             & 1.02\%                                  \\ \hline
all features                                              & \cellcolor[HTML]{9AFF99}0.58\%              & \cellcolor[HTML]{FFCE93}-0.97\%         & 0.33\%                                      & -0.51\%                                 & \cellcolor[HTML]{9AFF99}0.10\%              & 1.19\%                                  \\ \hline
\textsf{RIFS}                                                      & -2.70\%                                     & \cellcolor[HTML]{9AFF99}0.56\%          & 0.13\%                                      & \cellcolor[HTML]{9AFF99}1.83\%                                 & \cellcolor[HTML]{FFCE93}-1.70\%             & 2.45\%                                  \\ \hline
forward selection                                         & -0.13\%                                     & \cellcolor[HTML]{9AFF99}1.55\%          & \cellcolor[HTML]{FFCE93}-1.06\%             & -0.68\%                                 & \cellcolor[HTML]{9AFF99}6.70\%              & 0.62\%                                  \\ \hline
linear svc                                                & \cellcolor[HTML]{9AFF99}0.56\%              & \cellcolor[HTML]{FFCE93}-0.82\%         & \cellcolor[HTML]{FFCE93}-0.04\%             & -0.48\%                                 & \cellcolor[HTML]{FFCE93}-4.72\%             & \cellcolor[HTML]{FFCE93}-4.65\%         \\ \hline
relief                                                    & -0.01\%                                     & \cellcolor[HTML]{9AFF99}0.21\%          & 0.29\%                                     & \cellcolor[HTML]{FFCE93}-0.14\%         & \cellcolor[HTML]{FFCE93}-0.27\%             & 0.31\%                                  \\ \hline
\end{tabular}
\caption{Coreset construction sampling strategies for classification datasets: stratified sampling and sketching techniques (subspace embedding) described in \ref{subsection:coreset}. This table shows accuracy change of a given technique over uniform sampling.}
\label{tab:class_corest}
\end{table*}

\label{subsec:feature_selectors}
\begin{table}[htb!]
\small
\centering
\begin{tabular}{|c|
>{\columncolor[HTML]{FFCE93}}c |
>{\columncolor[HTML]{9aff99}}c |
>{\columncolor[HTML]{FFCE93}}c |}
\hline
\cellcolor[HTML]{EFEFEF}\textbf{Method} & \cellcolor[HTML]{EFEFEF}\textbf{Taxi} & \cellcolor[HTML]{EFEFEF}\textbf{Pickup} & \cellcolor[HTML]{EFEFEF}\textbf{Poverty} \\ \hline
\textsf{RIFS} & -0.37\% & 0.77\% & \cellcolor[HTML]{9aff99}0.01\% \\ \hline
sparse regression & \cellcolor[HTML]{9aff99}1.37\% & 0.03\% & \cellcolor[HTML]{9aff99}3.60\% \\ \hline
f-test & -3.42\% & 4.61\% & \cellcolor[HTML]{9aff99}0.00\% \\ \hline
lasso & \cellcolor[HTML]{9aff99}1.23\% & \cellcolor[HTML]{FFCE93}-0.50\% & -0.15\% \\ \hline
mutual info & \cellcolor[HTML]{9aff99}1.86\% & 0.13\% & -0.91\% \\ \hline
relief & -0.37\% & \cellcolor[HTML]{FFCE93}-4.21\% & -5.46\% \\ \hline
all features & \cellcolor[HTML]{9aff99}2.16\% & \cellcolor[HTML]{FFCE93}-0.08\% & -0.53\% \\ \hline
random forest & -2.25\% & \cellcolor[HTML]{FFCE93}-0.14\% & -6.21\% \\ \hline
forward selection & -2.06\% & \cellcolor[HTML]{FFCE93}-7.56\% & -3.07\% \\ \hline
\end{tabular}
\vspace{-0.1in}
\caption{Benchmarking sketching performance for various feature selection methods. The entries in the table indicate $\%$-change over uniform sampling.}
\label{tab:regress_corest}
\vspace{-0.2in}
\end{table}

\sparagraph{Coreset Construction.}
Next, we compare the performance of constructing coresets via Uniform sampling, Stratified sampling and Subspace Embedding. Recall, discussion for these sampling techniques can be found in Subsection \ref{subsection:coreset}. Results of the experiments are given in Table \ref{tab:regress_corest} and \ref{tab:class_corest}. 

The sketching algorithm (count-sketch) we use provably obtains a subspace embedding for the optimization objective in \textsf{RIFS} and Sparse Regression, given that the number of rows sampled are larger than the number of columns. While provable guarantees don't hold for our datasets (since columns are much larger) we expect sketching algorithms to perform well even with fewer rows.

We observe there is no one approach that always performs best, and the coreset performance is data and model dependent. \system{} uses uniform sampling as default sampling method, but allows users specify and explore other sampling strategies.

\sparagraph{Soft Joins on Time-Series.}
We then experiment with soft join techniques that we described in Section \ref{sec:joins}. In Figure \ref{fig:soft_join_types} we compare 4 different techniques for joining on a time series data: simple (hard) join on unmodified keys, Two-way Nearest Neighbour soft join, Nearest Neighbour soft join, and Time-Resampling technique for a simple (hard) join. Every Two-Way Nearest Neighbour and Nearest Neighbour joins also include time-resampling technique. 

We can see that in most cases Two-Way Nearest Neighbour soft join beats Nearest Neighbour and both outperform simple hard join. 
We address rounding concerns for hard join in the \textsf{Taxi} dataset by identifying a target time-granularity and aggregate rows of a foreign table that correspond to same days. 
We observe that \textsf{Taxi} dataset time-resampling technique performed much better combined with hard join (rather than soft). 

\begin{table}[htb!]
\small
\centering
\begin{tabular}{|c|c|c|c|c|}
\hline
\rowcolor[HTML]{EFEFEF} 
\textbf{Dataset} & \textbf{\begin{tabular}[c]{@{}c@{}}Score \\ change\end{tabular}} & \textbf{\begin{tabular}[c]{@{}c@{}}Speed  \\ (x faster)\end{tabular}} & \textbf{\begin{tabular}[c]{@{}c@{}}Tables \\ removed\end{tabular}} & \textbf{$\tau$} \\ \hline
Taxi             & -0.04\%                                                          & 3.18                                                                  & 10                                                                 & 24                           \\ \hline
Pickup           & -15.35\%                                                         & 3.50                                                                  & 17                                                                 & 17                           \\ \hline
Poverty          & -1.19\%                                                          & 5.87                                                                  & 36                                                                 & 15                           \\ \hline
School (S)       & -1\%                                                             & 1.14                                                                  & 2                                                                  & 15                           \\ \hline
School (L)       & -5\%                                                             & 1.32                                                                  & 39                                                                 & 17                           \\ \hline
\end{tabular}
\caption{Performance of ARDA with \textsf{RIFS} and Tuple Rule as a table filtering step for real world datasets. Hyperparameter $\tau$ was optimized for each dataset.}
\label{tab:TR_RIFS_prefilter}
\end{table}
\sparagraph{Tuple Ratio Test.}
Kumar et al. \cite{kumar2016join} propose several decision rules to safely eliminate tables before feature selection takes place. Here, we evaluate the most conservative rule proposed, the \textit{Tuple Ratio rule}. The \textit{Tuple Ratio} is defined as $\frac{n_S}{n_R}$, where $n_S$ is the number of training examples in a base table and ${n_R}$ is a size of a foreign-key domain. Based on an analysis of VC dimensions in binary classification problems, the decision rule suggests that a foreign table cannot help a predictive model if $\frac{n_S}{n_R}$ larger than a threshold. 

We note that this decision rule is intended only as a filtering technique (i.e., a small tuple ratio does not imply that a model will improve from the resulting join). We experimented with Tuple Ratio as a stand-alone method for feature augmentation and as a pre-filtering step. During the pre-filtering step we eliminate tables that are above optimized $\tau$ and therefore avoid feature selection. 
Kumar et al. suggest that $\tau$ only needs to be tuned per model, not per dataset. However, we found slight improvements from optimizing the threshold per dataset: we report the threshold used for each dataset in Table \ref{tab:TR_RIFS_prefilter}.


Tables \ref{tab:main_results} show that Tuple Ratio (shown as \textit{TR rule}) does not perform well as a stand-alone selection solution. Table \ref{tab:TR_RIFS_prefilter} shows experiments that use Tuple Ratio as a filtering tool to eliminate tables before feature selection takes place. We observed that filtering with the TR decision rule caused small to moderate decreases in model accuracy and medium to significant improvements in training time. 

The change in model accuracy is most pronounced on the Pickup dataset, which is not surprising given that the TR decision rule was designed for (binary) classification problems and the Pickup dataset represents a regression task. 
Overall, we conclude that the TR decision rule is a useful filtering technique for \system{} when training time is paramount and some model accuracy can be sacrificed. 

\sparagraph{Table grouping.}
Here we evaluate our three table grouping methods described in Section~\ref{sec:joins}: \emph{table-join}, \emph{budget-join}, and \emph{full materialization join}. Table~\ref{tab:join_plan_comparison} compares the performance of \emph{table-join}  and \emph{full materialization join} against \emph{budget-join} for various feature selectors. The fact that \emph{table-join} almost always performs worse than \emph{budget-join} is evidence that these datasets contain \emph{co-predictors}: some features are only useful when combined with other features from different tables. The presence of co-predictors also help explain why full materialization occasionally outperforms \emph{budget-join}. However, with RIFS, full materialization never outperforms \emph{budget-join} by a significant margin, and often degrades performance since full materialization results in many noise features which tend to interfere with the model's ability to learn. Since \emph{budget-join} is not significantly outperformed, we believe that it represents a reasonable compromise between full materialization and \emph{table-join} in terms of finding co-predictors (since \emph{budget-join} still considers multiple tables at once) and creating noise features.

\begin{table*}[]
\small
\centering
\begin{tabular}{|c|
>{\columncolor[HTML]{FFCE93}}c |
>{\columncolor[HTML]{9AFF99}}c |
>{\columncolor[HTML]{FFCE93}}c |c|
>{\columncolor[HTML]{FFCE93}}c |
>{\columncolor[HTML]{FFCE93}}c |
>{\columncolor[HTML]{FFCE93}}c |
>{\columncolor[HTML]{FFCE93}}c |}
\hline
\cellcolor[HTML]{EFEFEF} & \multicolumn{2}{c|}{\cellcolor[HTML]{EFEFEF}\textbf{Taxi}} & \multicolumn{2}{c|}{\cellcolor[HTML]{EFEFEF}\textbf{Pickup}} & \multicolumn{2}{c|}{\cellcolor[HTML]{EFEFEF}\textbf{Poverty}} & \multicolumn{2}{c|}{\cellcolor[HTML]{EFEFEF}\textbf{School(S)}} \\ \cline{2-9} 
\multirow{-2}{*}{\cellcolor[HTML]{EFEFEF}\textbf{Method}} & \cellcolor[HTML]{EFEFEF}\textbf{Table} & \cellcolor[HTML]{EFEFEF}\textbf{Fullmat} & \cellcolor[HTML]{EFEFEF}\textbf{Table} & \cellcolor[HTML]{EFEFEF}\textbf{Fullmat} & \cellcolor[HTML]{EFEFEF}\textbf{Table} & \cellcolor[HTML]{EFEFEF}\textbf{Fullmat} & \cellcolor[HTML]{EFEFEF}\textbf{Table} & \cellcolor[HTML]{EFEFEF}\textbf{Fullmat} \\ \hline
RIFS & {\color[HTML]{000000} -4.10\%} & {\color[HTML]{000000} 0.97\%} & {\color[HTML]{000000} -25.75\%} & \cellcolor[HTML]{9AFF99}{\color[HTML]{000000} 0.87\%} & {\color[HTML]{000000} -0.40\%} & {\color[HTML]{000000} -2.88\%} & {\color[HTML]{000000} -2.11\%} & {\color[HTML]{000000} -1.45\%} \\ \hline
forward selection & {\color[HTML]{000000} -3.92\%} & {\color[HTML]{000000} 1.11\%} & {\color[HTML]{000000} -6.60\%} & \cellcolor[HTML]{FFCE93}{\color[HTML]{000000} -7.53\%} & {\color[HTML]{000000} -8.66\%} & {\color[HTML]{000000} -0.50\%} & {\color[HTML]{000000} -1.00\%} & {\color[HTML]{000000} -0.14\%} \\ \hline
random forest & {\color[HTML]{000000} -3.55\%} & {\color[HTML]{000000} 2.86\%} & {\color[HTML]{000000} -22.01\%} & \cellcolor[HTML]{9AFF99}{\color[HTML]{000000} 0.04\%} & {\color[HTML]{000000} -7.92\%} & {\color[HTML]{000000} -3.20\%} & {\color[HTML]{000000} -1.85\%} & \cellcolor[HTML]{9AFF99}{\color[HTML]{000000} 0.12\%} \\ \hline
sparse regression & {\color[HTML]{000000} -5.35\%} & \cellcolor[HTML]{FFCE93}{\color[HTML]{000000} -3.27\%} & {\color[HTML]{000000} -25.30\%} & \cellcolor[HTML]{FFCE93}{\color[HTML]{000000} -12.67\%} & {\color[HTML]{000000} -2.51\%} & \cellcolor[HTML]{9AFF99}{\color[HTML]{000000} 0.92\%} & \cellcolor[HTML]{9AFF99}{\color[HTML]{000000} 0.44\%} & {\color[HTML]{000000} -0.29\%} \\ \hline
\end{tabular}
\caption{We compare change in final accuracy among several feature selectors with \emph{budget-join} being the baseline. \emph{table-join} performs one join at a time and \emph{full materialization} joins all tables simultaneously. 
}
\label{tab:join_plan_comparison}
\end{table*}

\sparagraph{Filtering Synthetic Noise.}
Using our micro benchmark datasets we compute the number of true and noisy features recovered by different feature selectors. We plot the resulting experiment in Figure \ref{fig:bar_selectors_comparisons}.

From Figure \ref{fig:bar_selectors_comparisons} we can conclude that the amount of noise that is being selected depends both on a feature selection method as well as on a data itself. While not perfect, \textsf{RIFS} shows best selectivity that filters out a lot of noise and redundant features from the original pool, while maintaining the top accuracy (see Table \ref{tab:microbenchs_main_results}).

\begin{table}[htb!]
\centering
\small
\begin{tabular}{|c|c|c|c|c|}
\hline
\rowcolor[HTML]{EFEFEF} 
\cellcolor[HTML]{EFEFEF} & \multicolumn{2}{c|}{\cellcolor[HTML]{EFEFEF}\textbf{Kraken}} & \multicolumn{2}{c|}{\cellcolor[HTML]{EFEFEF}\textbf{Digits}} \\ \cline{2-5} 
\rowcolor[HTML]{EFEFEF} 
\multirow{-2}{*}{\cellcolor[HTML]{EFEFEF}\textbf{Method}} & \textbf{acc.} & \textbf{time} & \textbf{acc.} & \textbf{time} \\ \hline
baseline (our) & 56.80\% & 7 & 39.77\% & 8 \\ \hline
all features (our) & 57.16\% & 29 & 90.97\% & 37 \\ \hline
\begin{tabular}[c]{@{}c@{}}all features\\ (Alpine Meadow)\end{tabular} & 52.72\% & 3600 & 81.56\% & 3600 \\ \hline
\begin{tabular}[c]{@{}c@{}}all features \\ (Azure AutoML)\end{tabular} & 79.63\% & 3600 & 92.14\% & 3600 \\ \hline
\begin{tabular}[c]{@{}c@{}}baseline \\ (Azure AutoML)\end{tabular} & 62.77\% & 3600 & 45.11\% & 3600 \\ \hline
\textsf{RIFS} & 71.44\% & 466 & 95.00\% & 172 \\ \hline
backward selection & 57.04\% & 4446 & 90.92\% & 1883 \\ \hline
forward selection & 64.54\% & 1527 & 94.24\% & 1118 \\ \hline
RFE & 57.18\% & 1059 & 94.07\% & 2174 \\ \hline
sparse regression & 62.76\% & 255 & 91.12\% & 397 \\ \hline
random forest & 65.42\% & 301 & 94.47\% & 147 \\ \hline
f-test & 74.20\% & 258 & 93.85\% & 213 \\ \hline
linear svc & 63.66\% & 188 & 91.08\% & 522 \\ \hline
logistic reg & 62.80\% & 156 & 91.38\% & 460 \\ \hline
mutual info & 57.46\% & 258 & 94.27\% & 227 \\ \hline
relief & 57.70\% & 302 & 91.29\% & 256 \\ \hline
\end{tabular}
\caption{Results on micro benchmark datasets on multiple feature selectors. \textsf{acc.} represents accuracy metric.}
\label{tab:microbenchs_main_results}
\end{table}

%% file: related_work.tex
\section{Related Work}
\label{sec:related_work}
There has been extensive prior work on data mining, knowledge discovery, data augmentation and feature selection.
We provide a brief overview of this literature. Perhaps the most pertinent work  is Kumar et. al.~\cite{kumar2016join} and Shah et al.~\cite{shah2017key}, which study when joining tables to the base table is unnecessary (i.e., is highly unlikely to improve a predictive model). The proposed decision rules (which come with theoretical bounds) allow practitioners to rule out joining with specific tables entirely. Both \cite{kumar2016join} and~\cite{shah2017key} are orthogonal to this work, as we are concerned with effectively augmenting a base table to improve model accuracy as opposed to determining which joins are potentially safe to avoid. In our experimental study, we demonstrate that Kumar et al.'s decision rules can be used as a prefiltering technique for Arda. 


\textbf{Data discovery}. Data discovery systems deal with sharing datasets, searching new datasets, and discovering relationships within heterogeneous data pools~\cite{bhardwaj2015collaborative, bhardwaj2014datahub, bhattacherjee2015principles, halevy2013data, gonzalez2010google, gatterbauer2006table, cafarella2009data, terrizzano2015data,halevy2016managing,halevy2016goods,terrizzano2015data, castro2017demo}. For example, Aurem~\cite{fernandez2018aurum} helps users automatically identify joins between tables representing similar entities. While these systems help users discover new data and explore relationships between datasets, they do not automatically determine whether or not such new information is useful for a predictive model. Arda uses data discovery systems as an input, combining new datasets and discovered relationships into more powerful predictive models.


\textbf{Data augmentation}. In general, data augmentation involves combining a dataset with additional information (derived or otherwise) in order to improve the accuracy of a predictive model. For example, learned embedding techniques~\cite{le2014distributed, deng2018table2vec} like Word2Vec~\cite{mikolov2013distributed} use a large corpus of unlabeled data to learn an embedding that can then be used to semantically represent new pieces of information. Other systems, like Octopus and InfoGather~\cite{yakout2012infogather} automatically search the web for additional pieces of information relevant to a user's data (e.g., discover the ZIP codes and populations of a list of cities).

\textbf{Feature selection.}
A natural approach to our problem would be to join all compatible tables and rely of the learning algorithm to ignore irrelevant features. The number of irrelevant features may be much greater than the number of relevant features, especially in large data lakes with high data diversity. 
Unfortunately, almost all ML algorithms are highly sensitive to noise, especially when the noise overwhelms the relevant features~\cite{madry2017towards,wong2017provable,szegedy2013intriguing,carlini2017adversarial,athalye2017synthesizing}. 

The negative impact of noise on model performance is well-studied~\cite{domingos1999role, bias_variance}, resulting in several works on feature selection, the task of narrowing down a large number of features to a smaller set of useful features. For an overview, see~\cite{kohavi1997wrappers}.

\textbf{\AML.} Recently, Automatic Machine Learning (\AML) has emerged as independent area of research seeking methods to automate the process of finding the best machine learning pipelines for a given learning task. \AML systems often handle ML tasks like feature selection, model selection, and hyperparameter tuning. Examples of \AML tools include Auto-sklearn \cite{feurer2015efficient}, Alpine Meadow \cite{10.1145/3299869.3319863}, Microsoft Azure AutoML \footnote{\url{azure.microsoft.com/en-us/services/machine-learning}}, Google Cloud AutoML \footnote{\url{cloud.google.com/automl}}, Neural Architecture Search \cite{bello2017neural, baker2016designing}, and others. However, these systems generally rely exclusively on the data supplied by the user and expect input in single-table form. While Arda can use any \AML system to construct a final model from the features Arda discovers, many techniques presented here could also be integrated into an \AML system.




%% file: conclusions.tex
\section{Conclusion \& Future Work}
We built a system that automates the data augmentation workflow and integrates it with the ML model selection process. We demonstrated the effectiveness and versatility of our system through a vast set of experiments that benchmark different available approaches for coreset construction, joining, feature selection etc. We addressed methods for joining on date-time keys, albeit location-based joins remain unexplored. Given the modular nature of our system, it would be interesting to evaluate sophisticated methods for data imputation, neural networks as learning models, and statistical significance tests for augmented features. While our work concentrated around achieving augmentation for a single base table, we hope future work can address automation of augmentation via transitive joins.